# JieHua Paintings Style Feature Extracting Model using Stable Diffusion with ControlNet

Yujia Gu[1]
California State University, Long Beach, United States, guyujia_2024@126.com

Li Haofeng[1]
Accademia di Belle Arti di Roma, Rome, Italy, lihaofeng_2024@126.com
Yujia Gu and Li Haofeng are first co-authors

Xinyu Fang
Columbia University, New York, United States, fangxinyu_2023@126.com

Zihan Peng
No.1 Central Primary School Shanghai Huangpu, Shanghai, P.R. China, richard.pengzh@gmail.com

Yinan Peng*
Shanghai Palmim Information Technology Ltd, Shanghai, China, kevin@palmim.com

Yujia Gu and Li Haofeng are first co-authors.

This study proposes a novel approach to extract stylistic features of Jiehua: the utilization of the Fine-tuned Stable Diffusion Model with ControlNet (FSDMC) to refine depiction techniques from artists' Jiehua. The training data for FSDMC is based on the open-source Jiehua artist's work collected from the Internet, which were subsequently manually constructed in the format of (Original Image, Canny Edge Features, Text Prompt). By employing the optimal hyperparameters identified in this paper, it was observed FSDMC outperforms CycleGAN, another mainstream style transfer model. FSDMC achieves FID of 3.27 on the dataset and also surpasses CycleGAN in terms of expert evaluation. This not only demonstrates the model's high effectiveness in extracting Jiehua's style features, but also preserves the original pre-trained semantic information. The findings of this study suggest that the application of FSDMC with appropriate hyperparameters can enhance the efficacy of the Stable Diffusion Model in the field of traditional art style migration tasks, particularly within the context of Jiehua.

**CCS Concepts**: CCS-> Computing methodologies -> Artificial intelligence -> Machine learning -> Ensemble methods

## 1 INTRODUCTION

Style transfer has become a popular research topic in the field of computer vision and image processing, with a particular focus on the style transfer of Chinese landscape painting using stable diffusion techniques. Several recent studies have explored different methods and models to achieve high-quality style transfer results, especially in the context of Chinese ancient painting styles. Wang et al. [1] introduced the GLStyleNet model, which combines global and local pyramid features to achieve exquisite style transfer with superior quality. Their approach has been demonstrated to be effective, particularly in tasks related to Chinese ancient painting style transfer. This highlights the importance of incorporating both global and local features in the style transfer process to enhance the overall quality of the results. On the other hand, Lee [2] proposed the Stable Style Transformer, which follows a delete and generate approach with an encoder-decoder architecture for text style transfer. While this work focuses on text style transfer, the two-stage approach could potentially be adapted for image style transfer tasks, including Chinese landscape painting styles. Gao et al. [3] shifted the focus from transferring styles of artistic paintings to the automatic generation of realistic paintings using generative adversarial networks (GANs). By leveraging advanced methods such as Neural Style Transfer (NST) and unsupervised cross-domain image translation, this study explores the generation of realistic paintings, including landscapes and portraits, through machine learning techniques. Liu [4] addressed the challenges in training algorithms for image oil painting style transfer by proposing an improved generative adversarial network based on gradient penalty. By incorporating total variance loss constraints, the study aimed to enhance the edge and texture details in the migration process of image oil painting styles. This highlights the importance of stability and convergence in the training process for achieving high-quality style transfer results. Peng et al. [5] introduced the Contour-enhanced CycleGAN Framework for style transfer from scenery photos to Chinese landscape paintings. By enhancing the contours in the style transfer process, the framework aimed to improve the fidelity and realism of the transferred images, particularly in the context of Chinese landscape painting styles. Moreover, Way et al. [6] proposed the TwinGAN model for imitating multiple styles of Chinese landscape paintings. By leveraging deep learning techniques, this study aimed to provide a comprehensive analysis of different Chinese landscape painting styles and compare them with existing works in the field of style transfer. In conclusion, the literature on style transfer of Chinese landscape painting using stable diffusion techniques showcases a diverse range of approaches and models aimed at achieving high-quality and realistic style transfer results. By incorporating global and local features, leveraging generative adversarial networks, and enhancing contours in the transfer process, researchers have made significant strides in advancing the field of style transfer for Chinese landscape painting styles. Further research in this area could focus on refining existing models, exploring new techniques, and expanding the application of style transfer to other artistic domains. These methods help to extract the feature embedding of Chinese landscape paintings more efficiently and completely, paving the way for further innovation in this field. Based on the previous research, this paper will aim to find out the generative model for extracting the features of Jiehua [7], which is a type of Chinese landscape paintings. Main contributions of this research include:

1) We collected datasets of ink paintings, oil paintings, and Jiehua on the Internet, including open source works by masters such as Zhang Xiao Gang, Yu Zhi Ding, and Yao Wen Han.
2) We used CycleGAN Models to Learn jiehua and Evaluate Its Style Transfer Ability
3) We leveraged Stable Diffusion Model with ControlNet to Learn Jiehua and Evaluate its Style Transfer Capability
4) Comparison of the Style Transfer Capabilities of the two models was carried out by both machine and manual evaluation.

## 2   RELATED WORK

**CycleGAN**: Unpaired Image-to-Image Translation using Cycle-Consistent Adversarial Networks has been a groundbreaking work in the field of image-to-image translation. Introduced by Zhu et al. [8], CycleGAN is capable of transforming images from one domain to another without the need for paired examples. This is achieved through the innovative use of cycle consistency loss, which ensures that the forward and reverse translations are consistent, thus preserving the original image content.

The key features of CycleGAN include its ability to handle unpaired data and its robust performance in various image translation tasks. It has been successfully applied in a wide range of applications, such as artistic style transfer, data augmentation for medical imaging, and cross-domain adaptation in computer vision tasks.

Compared to other GAN-based approaches, CycleGAN stands out for its flexibility and effectiveness in scenarios where paired training data is scarce or non-existent. However, it is important to note that CycleGAN may still face challenges in certain cases, such as handling fine-grained details or maintaining color consistency across translations.

In the context of related works, it is essential to acknowledge the contributions of other researchers who have built upon or extended the ideas of CycleGAN.

**Stable Diffusion Model**: The advent of generative models in the realm of deep learning has opened new avenues for creating realistic and diverse data samples. Among these, diffusion models have emerged as a powerful framework for generating high-quality images and other data types. This section reviews the evolution from latent diffusion models to the more refined Stable Diffusion Model.

Latent Diffusion Models are a class of generative models that operate on a latent space representation of data. They have been pivotal in advancing the field of image synthesis, offering a probabilistic approach to model the data distribution. The foundational work in this area, such as that by Rombach et al. [9], laid the groundwork for subsequent developments in the field.

Building upon the principles of latent diffusion, the Stable Diffusion Model introduces a series of innovations that enhance the stability and efficiency of the diffusion process. Unlike its predecessors, the Stable Diffusion Model, incorporates advanced techniques to stabilize the diffusion process, leading to faster convergence and improved sample quality.

The Stable Diffusion Model has found applications in a variety of domains, including but not limited to image generation, where it has demonstrated its ability to produce high-fidelity images with diverse content, and style transfer, where it can adapt the aesthetic of an image while preserving its semantic content.

**ControlNet**: In the realm of text-to-image synthesis, the advent of diffusion models has marked a significant leap forward, enabling the creation of visually striking images from textual descriptions However, these models often struggle with precise spatial control, which limits their ability to generate images that accurately reflect complex compositions and forms as envisioned by the user. To address this limitation, Zhang et al. introduce ControlNet, a neural network architecture designed to integrate spatial conditioning controls into large, pretrained text-to-image diffusion models [10].

The paper demonstrates the effectiveness of ControlNet in conditioning image generation using various inputs such as Canny edges, human poses, and segmentation maps. Notably, the training of ControlNets is shown to be robust across datasets of varying sizes, from less than 50k to over 1 million images, suggesting its versatility and resilience against overfitting and catastrophic forgetting — common challenges in finetuning large models with limited data [10].

## 3  METHODOLOGY

**CycleGAN** : CycleGANs are a special variant of traditional GANs. They can also create new data samples, but do so by transforming the input samples instead of creating them from scratch. In other words, they learn to transform data from two data sources. These data can be selected by the scientist or developer who provided the dataset for this algorithm. In the case where the two data sources were images of dogs and images of cats, the algorithm was able to effectively be able to convert images of cats to images of dogs and vice versa.

CycleGAN is a neural network that learns two data conversion functions between two domains. One of them is $G(x)$. It converts a given sample $x \in X$ into an element of domain $Y$. The second is $F(y)$, which converts a sample element $y \in Y$ into an element of domain $X$.

$$G: X \to Y$$

$$F = Y \to X$$

As shown above the entire architecture has two GANs, which form a CycleGAN. To learn $F$ and $G$, two conventional GANs are used. each GAN has a network of generators inside that learns how to transform the data as needed. the first generator of the GAN learns to compute $F$, and the second generator of the GAN learns to compute $G$. The first generator of the GAN learns to compute $F$, and the second generator of the GAN learns to compute $G$.

$$G: Generates\ \hat{y}\ from\ sample\ x$$
$$F: Generates\ \hat{x}\ from\ sample\ y$$

Below are the definitions of the generator functions $G$ and $F$:

$$D_x: Discrinminates\ y\ from\ G(x)$$
$$D_y: Discrinminates\ from\ F(y)$$

Furthermore, each generator is associated with a discriminator that learns to distinguish the actual data y from the synthetic data $G(x)$.

Thus, CycleGAN consists of two generators and two discriminators that learn the transformation functions F and G. This structure is shown in the following figure 1:

For each GAN network loss function, each GAN generator will learn its corresponding transform function ($F$ or $G$) by minimising the loss. The generator loss is calculated by measuring the difference between the generated data and the target data (e.g., the difference between a generated image of a cat compared to an image of a real cat). The greater the difference, the higher the penalty the generator will be subjected to.

The discriminator loss is also used to train the discriminator to be good at distinguishing real data from synthetic data.

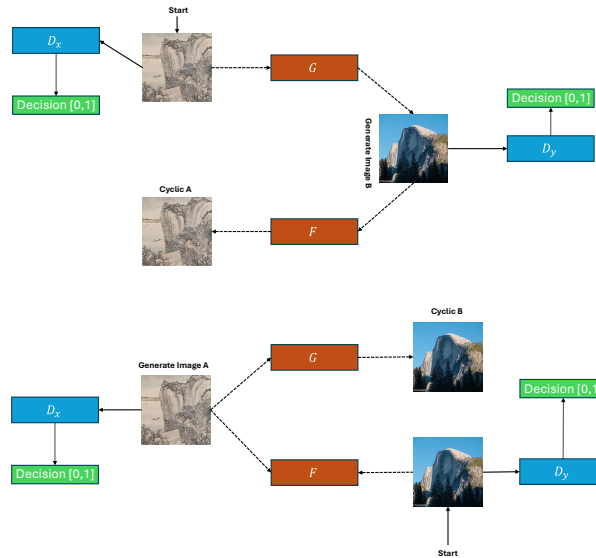

Figure 1. The architecture of CycleGAN for a Jiehua pair data

When these two are set together, they will improve each other. The generator is trained to spoof the discriminator, and the discriminator will be trained to better distinguish real data from synthetic data. As a result, the generator will be very good at creating/transforming the required data (learning the required transformations, e.g. F). Overall, GAN losses look like:

$$\text{Loss}_{GAN}(G, D_x) = E[\log(D_x(x))] + E[\log(1 - D_x(G(x)))]$$

As shown in the formula, the former $E[\log(D_x(x))]$ is the loss for Discriminator loss, while the latter is the loss for Generator loss.

For the second generator-discriminator pair, a similar loss can be written:
$$\text{Loss}_{GAN}(F, D_y) = E\left[\log\left(D_y(y)\right)\right] + E\left[\log\left(1 - D_y(F(y))\right)\right]$$

Training CycleGAN using only GAN losses is not guaranteed to maintain cycle consistency. Therefore, an additional cyclic consistency loss is used to enforce this property. Define this loss as the difference between the input value x and the forward prediction $F(G(x))$ and the input value y and the forward prediction $G(F(y))$. The larger the difference, the further away the prediction is from the original input. Therefore, the whole loss formula is like that:
$$\text{Loss} = \text{Loss}_{GAN}(F, D_y) + [E|G(F(x)) - x| + E|F(G(y)) - y|]$$

**Stable Diffusion with ControlNet**

In a traditional diffusion model without ControlNet, where the original neural network F input $x$ obtains $y$, the parameters are denoted by $\Theta$. And the mapping relationship is given in the following equation:
$$y = F(x; \Theta)$$

In ControlNet, it is the original neural network $F$ of the model that is locked, set as locked copy.

Then a copy of the original network is made, called a trainable copy, on which operations are performed to impose control conditions. The results of the imposed control conditions are then added to the results of the original model to obtain the final output.

After all these operations, after applying the control conditions, the output of the original network is finally modified to:
$$\boldsymbol{y}_c = F(\boldsymbol{x}; \Theta) + Z(F(\boldsymbol{x} + Z(\boldsymbol{c}; \Theta_{z1}); \Theta_c); \Theta_{z2})$$

Where zero convolution, or zero convolution layer $Z$ is initialized with weight and bias of 0. The parameters of the two zero convolution layers are $\{\Theta_{z1}, \Theta_{z2}\}$.

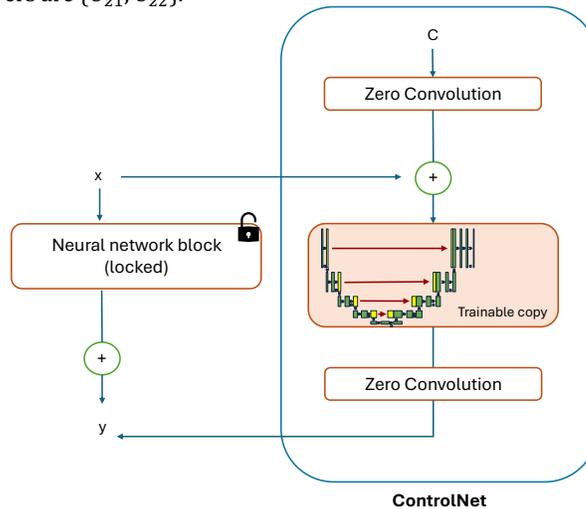

**ControlNet**

Figure 2. The architecture of Stable Diffusion with ControlNet

The structure of the model is shown in the figure above, where the control conditions are added to the original inputs after being zero-convolutionally added, and then added to the replicated neural network block of ControlNet (Trainable copy), where the output of the network is added to the output of the original network after doing zero-convolution once more.

The untrained ControlNet parameters after initialization should look like this:

$$\begin{cases} Z(c; \Theta_{z1}) = 0 \\ F(x + Z(c; \Theta_{z1}); \Theta_c) = F(x; \Theta_c) = F(x; \Theta) \\ Z(F(x + Z(c; \Theta_{z1}); \Theta_c); \Theta_{z2}) = Z(F(x; \Theta_c); \Theta_{z2}) = \mathbf{0} \end{cases}$$

This means that when ControlNet is untrained, the output is 0, and the number added to the original network is also 0. This has no effect on the original network and ensures that the performance of the original network is preserved. Afterwards, ControlNet training is only optimized on the original network, which can be considered the same as fine-tuning the network.

Then, for the loss function during training, the Stable Diffusion with ControlNet uses the following functions:

$$\mathcal{L} = \mathbb{E}_{z_0, t, c_t, c_r, \epsilon \sim \mathcal{N}(0,1)} \left[ \| \epsilon - \epsilon_\theta(z_t, t, c_t, c_f) \|_2^2 \right]$$

Where the latent variables obtained from the sampled $Z_t$ after denoising using the network $\epsilon_\theta$ and the original image through the network $\epsilon$ were used to calculate $L2 loss$ to see how well they were reconstructed.

The latent variable is obtained from the original image after $\epsilon$, and the graph after reconstruction by the network $\epsilon_\theta$ is calculated as $L2 loss$. The original Stable Diffusion decoder must deal with sample $z_t$ and time step $t$, where two control conditions have been added: the text prompt $C_t$ and the task-related prompt $C_f$.

During training 50 % of the textual cues $C_t$ were randomly replaced with empty strings. This facilitates the identification of semantic content from control conditions by the ControlNet network. The purpose of this is that when Stable Diffusion does not have a prompt, the encoder is able to get more semantics from the input control condition in place of the prompt, which is also known as classifier-free guidance.

## EXPERIMENT

- Dataset and Model Description

There are 112 images in the dataset used in this paper, with 44 Jiehuas and the other traditional paintings draw by Chinese artists. And all images are in the shape of (512,512). Then it is to construct the dataset for fine-tuning Stable Diffusion Model with ControlNet, here firstly the name of each artist is used as the text prompt for two kinds of Jiehua. with the prompt data, canny edge extraction is performed for each original image, and the result is used as the null condition image. Finally the (original image, control condition, text prompt) format is formed to fine tune the model.

Fot the Stable Diffusion Model with ControlNet, the pre-trained Text-to-Image model used in this paper is fine-tuned based on 'Stable-Diffusion-v1-5', which checkpoint was initialized with the weights of the Stable-Diffusion-v1-2 checkpoint.

Important hyperparameters during fine-tuning are plotted below:

Table 1: HYPERPARAMETERS DURING FINE-TUNING

| Hyperparameter | Value |
| --- | --- |
| learning_rate | 5e-6 |

| Hyperparameter | Value |
|---|---|
| lr_scheduler | cosine_with_restarts |
| lr_warmup_steps | 100 |
| gradient_accumulation_steps | 5 |
| resolution | 64 |
| use_ema | False |

And for the CycleGAN, we just set the default hyperparameters during training except for batch_size and learning_rate. Because of CycleGAN specifically designed to deal with small dataset size, this paper did not use data augmentation.

- Evaluation Metric

The present study employs the FID score as a means of conducting an objective evaluation of the results. Furthermore, three experts in the field of Jiehua were invited to provide an evaluation.

The FID score represents an improvement on the IS. It is generated by comparing the generated image with the real image and calculating the 'distance value', which is used to generate the evaluation score. A smaller indicator value indicates a better result. The FID score is generated by comparing the generated image with the real image and calculating the 'distance value', which is generated by comparing the generated image with the real image. A smaller value of the metric indicates a better result.

- Experiment result

In the experimental setup, the fine-tuned Stable Diffusion Model with ControlNet is used to generate an image based on the prompt 'Green grasslands, Grey sky, People in bright colours, {artist's name} style', and then a random picture in the training set is used for FID calculation. This process is repeated 10 times to get the average FID result. And for the CycleGAN after full training, this paper chose to put other non-Jiehua images in the dataset inside the model for inference. And then we used those predicted images to calculate the FID value with the same strategy as the former. The results are as follows:

Table 2. FID RESULTS FOR EACH MODELS

| Models | FID |
|---|---|
| CycleGAN | 56 |
| Fine-tuned Stable Diffusion with ControlNet | 3.27 |

From the above table, it can be clearly seen that the FID of CycleGAN is seriously higher than Fine-tuned Stable Diffusion with ControlNet. this means that CycleGAN is less capable of extracting the features of Jiehua. And switching to other style migration images, such as Zebra to Horse, re-trained CycleGAN again shows amazing performance. Therefore, in addition to the reason of data volume, CycleGAN is worth exploring in Jiehua feature lifting ability.

Similarly, Fine-tuned Stable Diffusion with ControlNet is also ahead of CycleGAN in the expert scoring session.

The following figure 3 is the inference result based on Fine-tuned Stable Diffusion with ControlNet, and it can be clearly seen that the ability of the model to extract Jiehua after fine-tuning is very strong.

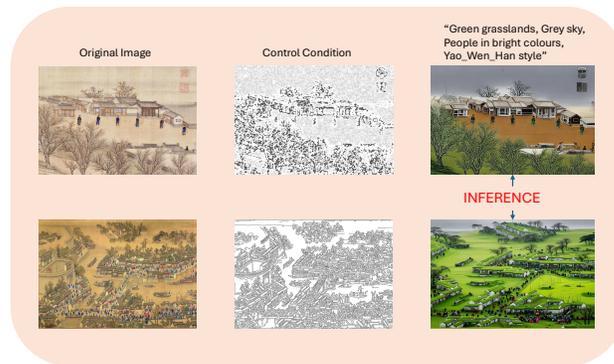

Figure 3. The inference results with Fine-tuned Stable Diffusion Model with ControlNet

## 4 CONCLUSION

In this paper, we investigate the performance of Fine-tuned Stable Diffusion Model with ControlNet (FSDMC) and CycleGAN, both mainstream models for Style Transfer tasks, in getting the style features of Jiehua painting. Firstly, in the introduction section, we outline some of the Generative models and its application in Style Transfer task. Subsequently, in the Related Work section, we introduce CycleGAN and Stable Diffusion Model with ControlNet and use them as comparative models in the Experimental section. In the methodology section, we present the essential principles of two completely distinct models. In the experimental part, we compare the 2 models with FID evaluation metric and experts rating, which conclude that the FSDMC has a very powerful performance on Jiehua Paintings.